%% file: face_recognition.tex
\documentclass[conference]{IEEEtran}

\usepackage{graphicx}
\usepackage{amsmath}
\usepackage{bm}
\usepackage{amsfonts}
\usepackage{mathtools}
\usepackage[utf8]{inputenc}
\usepackage[hidelinks]{hyperref}
\usepackage{subcaption}
\usepackage[capitalize,noabbrev]{cleveref}
\usepackage{algorithm}
\usepackage{algpseudocode}
\usepackage{booktabs}
\usepackage{notoccite}
\usepackage{fancyhdr}
\usepackage{diagbox}
\usepackage{tabularx}

\DeclareMathOperator*{\argmax}{arg\,max}

\DeclarePairedDelimiter\norm{\lVert}{\rVert}

\creflabelformat{equation}{#2\textup{#1}#3}

\begin{document}

\title{Face Recognition: From Traditional to Deep Learning Methods}

\author{
\IEEEauthorblockN{Daniel Sáez Trigueros, Li Meng}
\IEEEauthorblockA{School of Engineering and Technology\\
University of Hertfordshire\\
Hatfield AL10 9AB, UK}
\and
\IEEEauthorblockN{Margaret Hartnett}
\IEEEauthorblockA{GBG plc\\
London E14 9QD, UK}
}

\maketitle

\begin{abstract}
Starting in the seventies, face recognition has become one of the most researched topics in computer vision and biometrics. Traditional methods based on hand-crafted features and traditional machine learning techniques have recently been superseded by deep neural networks trained with very large datasets. In this paper we provide a comprehensive and up-to-date literature review of popular face recognition methods including both traditional (geometry-based, holistic, feature-based and hybrid methods) and deep learning methods.
\end{abstract}

\section{Introduction}
\label{sec:introduction}

Face recognition refers to the technology capable of identifying or verifying the identity of subjects in images or videos. The first face recognition algorithms were developed in the early seventies \cite{kelly1970visual,kanade1973picture}. Since then, their accuracy has improved to the point that nowadays face recognition is often preferred over other biometric modalities that have traditionally been considered more robust, such as fingerprint or iris recognition \cite{delac2004survey}. One of the differential factors that make face recognition more appealing than other biometric modalities is its non-intrusive nature. For example, fingerprint recognition requires users to place a finger in a sensor, iris recognition requires users to get significantly close to a camera, and speaker recognition requires users to speak out loud. In contrast, modern face recognition systems only require users to be within the field of view of a camera (provided that they are within a reasonable distance from the camera). This makes face recognition the most user friendly biometric modality. It also means that the range of potential applications of face recognition is wider, as it can be deployed in environments where the users are not expected to cooperate with the system, such as in surveillance systems. Other common applications of face recognition include access control, fraud detection, identity verification and social media.

Face recognition is one of the most challenging biometric modalities when deployed in unconstrained environments due to the high variability that face images present in the real world (these type of face images are commonly referred to as faces in-the-wild). Some of these variations include head poses, aging, occlusions, illumination conditions, and facial expressions. Examples of these are shown in \cref{fig:face_variations}.

\input{figures/fig_face_variations}

\input{figures/fig_face_recognition_blocks}

Face recognition techniques have shifted significantly over the years. Traditional methods relied on hand-crafted features, such as edges and texture descriptors, combined with machine learning techniques, such as principal component analysis, linear discriminant analysis or support vector machines. The difficulty of engineering features that were robust to the different variations encountered in unconstrained environments made researchers focus on specialised methods for each type of variation, e.g. age-invariant methods \cite{park2010age,li2011discriminative}, pose-invariant methods \cite{ding2016comprehensive}, illumination-invariant methods \cite{liu2005illumination,tan2010enhanced}, etc. Recently, traditional face recognition methods have been superseded by deep learning methods based on convolutional neural networks (CNNs). The main advantage of deep learning methods is that they can be trained with very large datasets to learn the best features to represent the data. The availability of faces in-the-wild on the web has allowed the collection of large-scale datasets of faces \cite{sun2014deep,yi2014learning,parkhi2015deep,guo2016ms,nech2017level,bansal2017umdfaces,cao2017vggface2} containing real-world variations. CNN-based face recognition methods trained with these datasets have achieved very high accuracy as they are able to learn features that are robust to the real-world variations present in the face images used during training. Moreover, the rise in popularity of deep learning methods for computer vision has accelerated face recognition research, as CNNs are being used to solve many other computer vision tasks, such as object detection and recognition, segmentation, optical character recognition, facial expression analysis, age estimation, etc.

Face recognition systems are usually composed of the following building blocks:

\begin{enumerate}
    \item \textbf{Face detection}. A face detector finds the position of the faces in an image and (if any) returns the coordinates of a bounding box for each one of them. This is illustrated in \cref{fig:face_detection_alignment_a}.
    \item \textbf{Face alignment}. The goal of face alignment is to scale and crop face images in the same way using a set of reference points located at fixed locations in the image. This process typically requires finding a set of facial landmarks using a landmark detector and, in the case of a simple 2D alignment, finding the best affine transformation that fits the reference points. \cref{fig:face_detection_alignment_b,fig:face_detection_alignment_c} show two face images aligned using the same set of reference points. More complex 3D alignment algorithms (e.g. \cite{hassner2015effective}) can also achieve face frontalisation, i.e. changing the pose of a face to frontal.
    \item \textbf{Face representation}. At the face representation stage, the pixel values of a face image are transformed into a compact and discriminative feature vector, also known as a template. Ideally, all the faces of a same subject should map to similar feature vectors.
    \item \textbf{Face matching}. In the face matching building block, two templates are compared to produce a similarity score that indicates the likelihood that they belong to the same subject.
\end{enumerate}

Face representation is arguably the most important component of a face recognition system and the focus of the literature review in \cref{sec:literature_review}.

\input{figures/fig_face_detection_alignment}

\section{Literature Review}
\label{sec:literature_review}

Early research on face recognition focused on methods that used image processing techniques to match simple features describing the geometry of the faces. Even though these methods only worked under very constrained settings, they showed that it is possible to use computers to automatically recognise faces. After that, statistical subspaces methods such as principal component analysis (PCA) and linear discriminant analysis (LDA) gained popularity. These methods are referred to as \textit{holistic} since they use the entire face region as an input. At the same time, progress in other computer vision domains led to the development of local feature extractors that are able to describe the texture of an image at different locations. \textit{Feature-based} approaches to face recognition consist of matching these local features across face images. Holistic and feature-based methods were further developed and combined into \textit{hybrid} methods. Face recognition systems based on hybrid methods remained the state-of-the-art until recently, when deep learning emerged as the leading approach to most computer vision applications, including face recognition. The rest of this paper provides a summary of some of the most representative research works on each of the aforementioned types of methods.

\subsection{Geometry-based Methods}
\label{geometry-based_methods}

Kelly's \cite{kelly1970visual} and Kanade's \cite{kanade1973picture} PhD theses in the early seventies are considered the first research works on automatic face recognition. They proposed the use of specialised edge and contour detectors to find the location of a set of facial landmarks and to measure relative positions and distances between them. The accuracy of these early systems was demonstrated on very small databases of faces (a database of 10 subjects was used in \cite{kelly1970visual} and a database of 20 subjects was used in \cite{kanade1973picture}). In \cite{brunelli1993face}, a geometry-based method similar to \cite{kanade1973picture} was compared with a method that represents face images as gradient images. The authors showed that comparing gradient images provided better recognition accuracy than comparing geometry-based features. However, the geometry-based method was faster and needed less memory. The feasibility of using facial landmarks and their geometry for face recognition was thoroughly studied in \cite{shi2006effective}. Specifically, they proposed a method based on measuring the Procrustes distance \cite{dryden1998statistical} between two sets of facial landmarks and a method based on measuring ratios of distances between facial landmarks. The authors argued that even though other methods that extract more information from the face (e.g. holistic methods) could achieve greater recognition accuracy, the proposed geometry-based methods were faster and could be used in combination with other methods to develop hybrid methods. Geometry-based methods have proven more effective in 3D face recognition thanks to the depth information encoded in 3D landmarks \cite{daniyal2009compact,gupta2010anthropometric}.

Geometry-based methods were crucial during the early days of face recognition research. They can be used as a fast alternative to (or in combination with) the more advanced methods described in the rest of this review.

\subsection{Holistic Methods}
\label{holistic_methods}

Holistic methods represent faces using the entire face region. Many of these methods work by projecting face images onto a low-dimensional space that discards superfluous details and variations not needed for the recognition task. One of the most popular approaches in this category is based on PCA. The idea, first proposed in \cite{sirovich1987low,kirby1990application}, is to apply PCA to a set of training face images in order to find the eigenvectors that account for the most variance in the data distribution. In this context, the eigenvectors are typically called \textit{eigenfaces} due to their resemblance to real faces, as shown in \cref{fig:eigenfaces}. New faces can be projected onto the subspace spanned by the eigenfaces to obtain the weights of the linear combination of eigenfaces needed to reconstruct them. This idea was used in \cite{turk1991eigenfaces} to identify faces by comparing the weights of new faces to the weights of faces in a gallery set. A probabilistic version of this approach based on a Bayesian analysis of image differences was proposed in \cite{moghaddam1998beyond}. In this method, two sets of eigenfaces were used to model intra-personal and inter-personal variations separately. Many other variations of the original eigenfaces method have been proposed. For example,
a nonlinear extension of PCA based on kernel methods, namely kernel PCA \cite{scholkopf1997kernel}, was proposed in \cite{kim2002face}; independent component analysis (ICA) \cite{comon1994independent}, a generalisation of PCA that can capture high-order dependencies between pixels, was proposed in \cite{bartlett2001independent}; and a two-dimensional PCA based on 2D image matrices instead of 1D vectors was proposed in \cite{yang2004two}.

\input{figures/fig_eigenfaces}

One issue with PCA-based approaches is that the projection maximises the variance across all the images in the training set. This implies that the top eigenvectors might have a negative impact on the recognition accuracy since they might correspond to intra-personal variations that are not relevant for the recognition task (e.g. illumination, pose or expression). Holistic methods based on \textit{linear discriminant analysis} (LDA), also called \textit{Fisher discriminant analysis}, \cite{fisher1938statistical} have been proposed to solve this issue \cite{belhumeur1997eigenfaces,etemad1997discriminant,zhao1998discriminant,zhao1999subspace}. The main idea behind LDA is to use the class labels to find a projection matrix $\bm{W}$ that maximises the variance between classes while minimising the variance within classes:
\input{equations/eq_lda_projection_matrix}
where $\bm{S}_w$ and $\bm{S}_b$ are the between-class and within-class scatter matrices defined as follows:
\input{equations/eq_lda_scatter_matrices}
where $\bm{x}_j$ represents a data sample, $\bm{\mu}_k$ is the mean of class $C_k$, $\bm{\mu}$ is the overall mean and $K$ is the number of classes in the dataset. The solution to \cref{eq:lda_projection_matrix} can be found by computing the eigenvectors of the separation matrix $\bm{S} = \bm{S}_w^{-1}\bm{S}_b$. Similar to PCA, LDA can be used for dimensionality reduction by selecting a subset of eigenvectors corresponding to the largest eigenvalues. Even though LDA is considered a more suitable technique for face recognition than PCA, pure LDA-based methods are prone to overfitting when the within-class scatter matrix $\bm{S}_w$ is not correctly estimated \cite{zhao1998discriminant,zhao1999subspace}. This happens when the input data is high-dimensional and not many samples per class are available during training. In the extreme case, $\bm{S}_w$ becomes singular and $\bm{W}$ cannot be computed \cite{belhumeur1997eigenfaces}. For this reason, it is common to reduce the dimensionality of the data with PCA before applying LDA \cite{belhumeur1997eigenfaces,zhao1998discriminant,zhao1999subspace}. LDA has also been extended to the nonlinear case using kernels \cite{mika1999fisher,liu2002face} and to probabilistic LDA \cite{ioffe2006probabilistic}.

\textit{Support vector machines} (SVMs) have also been used as holistic methods for face recognition. In \cite{phillips1999support}, the task was formulated as a two-class problem by training an SVM with image differences. More specifically, the two classes are the within-class difference set, which contains all the differences between images of the same class, and the between-class difference set, which contains all the differences between images of distinct classes (this formulation is similar to the probabilistic PCA proposed in \cite{moghaddam1998beyond}). In addition, \cite{phillips1999support} modified the traditional SVM formulation by adding a parameter to control the operating point of the system. In \cite{jonsson2002support}, a separate SVM was trained for each class. The authors experimented with SVMs trained with PCA projections and with LDA projections. It was found that this SVM approach only gives better performance compared with simple Euclidean distance when trained with PCA projections, since LDA already encodes the discriminant information needed to recognise faces.

An approach related to PCA and LDA is the \textit{locality preserving projections} (LPP) method proposed in \cite{he2005face}. While PCA and LDA preserve the global structure of the image space (maximising variance and discriminant information respectively), LPP aims to preserve the local structure of the image space. This means that the projection learnt by LPP maps images with similar local information to neighbouring points in the LPP subspace. For example, two images of the same person with open and closed mouth would be mapped to similar points using LPP, but not necessarily with PCA or LDA. This approach was shown to be superior than PCA and LDA on multiple datasets. Further improvements were achieved in \cite{cai2006orthogonal} by
making the LPP basis vectors orthogonal.

Another popular family of holistic methods is based on sparse representation of faces. The idea, first proposed in \cite{wright2009robust} as \textit{sparse representation-based classification} (SRC), is to represent faces using a linear combination of training images:
\input{equations/eq_src_representation}
where $\bm{y}$ is a test image, $\bm{A}$ is a matrix containing all the training images and $\bm{x}_0$ is a vector of sparse coefficients. By enforcing sparsity in the representation, most of the nonzero coefficients belong to training images from the correct class. At test time, the coefficients belonging to each class are used to reconstruct the image, and the class that achieves the lowest reconstruction error is considered the correct one. The robustness of this approach to image corruptions like noise or occlusions can be increased by adding a term of sparse error coefficients $\bm{e}_0$ to the linear combination:
\input{equations/eq_src_representation_corrupted}
where the nonzero entries of $\bm{e}_0$ correspond to corrupted pixels. Many variations of this approach have been proposed for increased robustness and reduced computational complexity. For example, the discriminative K-SVD algorithm was proposed in \cite{zhang2010discriminative} to select a more compact and discriminative set of training images to reconstruct the images; in \cite{zhou2009face}, SRC was extended by using a Markov random field to model the prior assumption about the spatial continuity of the occluded regions; and in \cite{jia2008face}, it was proposed to weight each pixel in the image independently to achieve better reconstructed images.

More recently, inspired by probabilistic PCA \cite{moghaddam1998beyond}, the \textit{joint Bayesian} method \cite{chen2012bayesian} has been proposed. In this method, instead of using image differences as in \cite{moghaddam1998beyond}, a face image is represented as the sum of two independent Gaussian variables representing intra-personal and inter-personal variations. Using this method, an accuracy of 92.4\% was achieved on the challenging Labeled Faces in the Wild (LFW) dataset \cite{huang2007labeled}. This is the highest accuracy reported by a holistic method on this dataset.

Holistic methods have been of paramount importance to the development of real-world face recognition systems, as evidenced by the large number of approaches proposed in the literature. In the next subsection, a popular family of methods that evolved as an alternative to holistic methods, namely feature-based methods, is discussed.

\subsection{Feature-based Methods}
\label{feature-based_methods}

Feature-based methods refer to methods that leverage local features extracted at different locations in a face image. Unlike geometry-based methods, feature-based methods focus on extracting discriminative features rather than computing their geometry\footnote{Technically, geometry-based methods can be seen as a special case of feature-based methods, since many feature-based methods also leverage the geometry of the extracted features.}. Feature-based methods tend to be more robust than holistic methods when dealing with faces presenting local variations (e.g. facial expression or illumination). For example, consider two face images of the same subject in which the only difference between them is that the person's eyes are closed in one of them. In a feature-based method, only the coefficients of the feature vectors that correspond to features extracted around the eyes will differ between the two images. On the other hand, in a holistic method, all the coefficients of the feature vectors might differ. Moreover, many of the descriptors used in feature-based methods are designed to be invariant to different variations (e.g. scaling, rotation or translation).

One of the first feature-based methods was the modular eigenfaces method proposed in \cite{pentland1994view}, an extension of the original eigenfaces technique. In this method, PCA was independently applied to different local regions in the face image to produce sets of \textit{eigenfeatures}. Although \cite{pentland1994view} showed that both eigenfeatures and eigenfaces can achieve the same accuracy, the eigenfeatures approach provided better accuracy when only a few eigenvectors were used.

A feature-based method that uses binary edge features was proposed in \cite{takacs1998comparing}. Their main contribution was to improve the Hausdorff distance that was used in \cite{huttenlocher1993comparing} to compare binary images. The Hausdorff distance measures proximity between two set of points by looking at the greatest distance from a point in one set to the closest point in the other set. In the modified Hausdorff distance proposed in \cite{takacs1998comparing}, each point in one set has to be near some point in the other set. It was argued that this property makes the method more robust to small, non-rigid local distortions. A variation of this method proposed \textit{line edge maps} (LEMs) for face representation \cite{gao2002face}. LEMs provide a compact face representation since edges are encoded as line segments, i.e. only the coordinates of the end points are used. A line segment Hausdorff distance was also proposed in this work to match LEMs. The proposed distance is discouraged to match lines with different orientations, is robust to line displacements, and incorporates a measure of the difference between the number of lines found in two LEMs.

A very popular feature-based method was the \textit{elastic bunch graph matching} (EBGM) method \cite{wiskott1997face}, an extension of the \textit{dynamic link architecture} proposed in \cite{lades1993distortion}. In this method, a face is represented using a graph of nodes. The nodes contain Gabor wavelet coefficients \cite{lee1996image} extracted around a set of predefined facial landmarks. During training, a \textit{face bunch graph} (FBG) model is created by stacking the manually located nodes of each training image. When a test face image is presented, a new graph is created and fitted to the facial landmarks by searching for the most similar nodes in the FBG model. Two images can be compared by measuring the similarity between their graph nodes. A version of this method that uses histograms of oriented gradients (HOG) \cite{freeman1995orientation,dalal2005histograms} instead of Gabor wavelet features was proposed in \cite{albiol2008face}. This method outperforms the original EBGM method thanks to the increased robustness of HOG descriptors to changes in illumination, rotation and small displacements.

With the development of local feature descriptors in other computer vision applications \cite{mikolajczyk2005performance}, the popularity of feature-based methods for face recognition increased. In \cite{ahonen2006face}, histograms of LBP descriptors were extracted from local regions independently, as shown in \cref{fig:lbp_representation}, and concatenated to form a global feature vector. Additionally, they measured the similarity between two feature vectors $\bm{a}$ and $\bm{b}$ using a weighted Chi-square distance:
\input{equations/eq_chi_square_distance}
where $w_i$ is a weight that controls the contribution of the $i$-th coefficient of the feature vectors. As shown in \cite{huang2011local}, many variations of this method have been proposed to improve face recognition accuracy and to tackle other related tasks such as face detection, facial expression analysis and demographic classification. For example, LBP descriptors extracted from Gabor feature maps, known as LGBP descriptors, were proposed in \cite{zhang2005multi,zhang2005local}; a rotation invariant LBP descriptor that applies Fourier transform to LBP histograms was proposed in \cite{ahonen2009rotation}; and a variation of LBP called \textit{local derivative pattern} (LDP) was proposed in \cite{zhang2010local} to extract high-order local information by encoding directional pattern features.

\input{figures/fig_lbp_representation}

\textit{Scale-invariant feature transform} (SIFT) descriptors \cite{lowe1999object} have also been extensively used for face recognition. Three different methodologies for matching SIFT descriptors across face images were proposed in \cite{bicego2006use}: (i) computing the distances between all pairs of SIFT descriptors and using the minimum distance as a similarity score; (ii) similar to (i) but SIFT descriptors around the eyes and the mouth are compared independently, and the average of the two minimum distances is used as a similarity score; and (iii) computing SIFT descriptors over a regular grid and using the average distance between the corresponding pairs of descriptors as a similarity score. The best recognition accuracy was obtained using the third method. A related method \cite{dreuw2009surf} proposed the use of \textit{speeded up robust features} (SURF) \cite{bay2006surf} features instead of SIFT. In this work, the authors observed that dense feature extraction over a regular grid provides the best results. In \cite{geng2009face}, two variations of SIFT were proposed, namely, the \textit{volume-SIFT} which removes unreliable keypoints based on their scale, and the \textit{partial-descriptor-SIFT} which finds keypoints at large scales and near face boundaries. Compared to the original SIFT, both approaches were shown to improve face recognition accuracy.

Some feature-based methods have focused on learning local features from training samples. For example, in \cite{cao2010face}, unsupervised learning techniques (K-means \cite{lloyd1982least}, PCA tree \cite{freund2008learning} and random-projection tree \cite{freund2008learning}) were used to encode local microstructures of faces into a set of discrete codes. The discrete codes were then grouped into histograms at different facial regions. The final local descriptors were computed by applying PCA to each histogram. A learning-based descriptor with similarities to LBP was proposed in \cite{sharma2012local}. Specifically, this descriptor consists of a differential pattern generated by subtracting the centre pixel of a local $3\times3$ region to its neighbouring pixels and a training of a Gaussian mixture model to compute high-order statistics of the differential pattern. Another LBP-like descriptor that has a learning stage was proposed in \cite{lei2014learning}. In this work, LDA was used to (i) learn a filter that when applied to an image enhances the discriminative ability of the differential patterns, and (ii) learn a set of weights that are assigned to the neighbouring pixels within each local region to reflect their contribution to the differential pattern.

Feature-based methods have been shown to provide more robustness to different types of variations than holistic methods. However, some of the advantages of holistic methods are lost (e.g. discarding non-discriminant information and more compact representations). Hybrid methods that combine both of these approaches are discussed next.


\subsection{Hybrid Methods}
\label{hybrid_methods}

Hybrid methods combine techniques from holistic and feature-based methods. Before deep learning became widespread, most state-of-the-art face recognition systems were based on hybrid methods. Some hybrid methods simply combine two different techniques without any interaction between them. For example, in the modular eigenfaces work \cite{pentland1994view} covered earlier, the authors experimented with a combined representation using both eigenfaces and eigenfeatures and achieved better accuracy than using either of these two methods alone. However, the most popular hybrid approach is to extract local features (e.g. LBP, SIFT) and project them onto a lower-dimensional and discriminative subspace (e.g. using PCA or LDA) as shown in \cref{fig:hybrid_representation}.

Several hybrid methods that use Gabor wavelet features combined with different subspaces methods have been proposed \cite{liu2002gabor,liu2003independent,liu2004gabor}. In these methods, Gabor kernels of different orientations and scales are convolved with an image and their outputs are concatenated into a feature vector. The feature vector is then downsampled to reduce its dimensionality. In \cite{liu2002gabor}, the feature vector was further processed using the enhanced linear discriminant model proposed in \cite{liu2000robust}. PCA followed by ICA were applied to the downsampled feature vector in \cite{liu2003independent}, and the probabilistic reasoning model from \cite{liu2000robust} was used to classify whether two images belong to the same subject. In \cite{liu2004gabor}, kernel PCA with polynomial kernels was applied to the feature vector to encode high-order statistics. All these hybrid methods were shown to provide better accuracy than using Gabor wavelet features alone.

LBP descriptors have been a key component in many hybrid methods. In \cite{chan2007multi}, an image was divided into non-overlapping regions and LBP descriptors were extracted at multiple resolutions. The LBP coefficients at each region were concatenated into regional feature vectors and projected onto PCA+LDA subspaces. This approach was extended to colour images in \cite{chan2007multispectral}. Laplacian PCA, an extension of PCA, was shown to outperform standard PCA and kernel PCA when applied to LBP descriptors in \cite{zhao2007laplacian}. Two novel patch versions of LBP, namely three-patch LBP (TPLBP) and four-patch LBP (FPLBP), were combined with LDA and SVMs in \cite{wolf2008descriptor}. The proposed TPLBP and FPLBP descriptors can boost face recognition accuracy by encoding similarities between neighbouring patches of pixels. More recently, \cite{chen2013blessing} proposed a high-dimensional face representation by densely extracting multi-scale LBP (MLBP) descriptors around facial landmarks. The high-dimensional feature vector (100K-dim) was reduced to 400 dimensions by PCA and a final discriminative feature vector was learnt using joint Bayesian. In their experiments, \cite{chen2013blessing} showed that extracting high-dimensional features can increase face recognition accuracy by 6-7\% when going from 1K to 100K dimensions. The main drawback of this approach is the high computational costs needed to perform a dimensionality reduction of such magnitude. For this reason, they proposed to approximate the PCA and joint Bayesian transformations with a sparse linear projection matrix $\bm{B}$ by solving the following optimisation problem:
\input{equations/eq_sparse_projection}
where the first term is a reconstruction error between the matrix $\bm{X}$ of high-dimensional feature vectors and the matrix $\bm{Y}$ of projected low-dimensional feature vectors; the second term enforces sparsity in the projection matrix $\bm{B}$; and $\lambda$ balances the contribution of each term. Another recent method proposed a multi-task learning approach based on a discriminative Gaussian process latent variable model, named \textit{GaussianFace} \cite{lu2015surpassing}. This method extended the Gaussian process approach proposed in \cite{urtasun2007discriminative} and incorporated a computationally more efficient version of kernel LDA to learn a face representation from LBP descriptors that can exploit data from multiple source domains. Using this method, an accuracy of 98.52\% was achieved on the LFW dataset. This is competitive with the accuracy achieved by many deep learning methods.

\input{figures/fig_hybrid_representation}

Some hybrid methods have proposed to use a combination of different local features. For example, Gabor wavelet and LBP features were used in \cite{tan2007fusing}. The authors argued that these two types of features capture complementary information. While LBP descriptors capture small appearance details, Gabor wavelet features encode facial shape over a broader range of scales. PCA was applied independently to the feature vectors containing the Gabor wavelet coefficients and the LBP coefficients to reduce their dimensionality. The final face representation was obtained by concatenating the two PCA-transformed feature vectors and applying a subspace method similar to kernel LDA, namely \textit{kernel discriminative common vector} \cite{cevikalp2006discriminative}. Another method that uses Gabor wavelet and LBP features was proposed in \cite{shan2006ensemble}. In this method, faces were represented by applying PCA+LDA to regions containing histograms of LGBP descriptors \cite{zhang2005local}. A multi-feature system was proposed in \cite{tan2010enhanced} to tackle face recognition under difficult illumination conditions. Three contributions were made in this work: (i) a preprocessing pipeline that reduces the effect of illumination variation; (ii) an extension of LBP, called \textit{local ternary patterns} (LTP), which is more discriminant and less sensitive to noise in uniform regions; and (iii) an architecture that combines sets of Gabor wavelet and LBP/LTP features followed by kernel LDA, score normalisation and score fusion. A related method \cite{chan2013multiscale} proposed a novel descriptor robust to blur that extends \textit{local phase quantization} (LPQ) descriptors \cite{rahtu2012local} to multiple scales (MLPQ). In addition, a kernel fusion technique was used to combine MLPQ descriptors with MLBP descriptors in the kernel LDA framework. In \cite{li2011discriminative}, an age invariant face recognition system was proposed based on dense extraction of SIFT and multi-scale LBP descriptors combined with a novel multi-feature discriminant analysis (MFDA). The MFDA technique uses random subspace sampling \cite{ho1998random} to construct multiple lower-dimensional feature subspaces, and bagging \cite{breiman1996bagging} to select subsets of training samples for LDA that contain inter-class pairs near the classification boundary to increase the discriminative ability of the representation. Dense SIFT descriptors were also used in \cite{saez2016shape} as texture features, and combined with shape features in the form of relative distances between pairs of facial landmarks. This combination of shape and texture features was further processed using multiple PCA+LDA transformations.

To conclude this subsection, other types of hybrids methods that do not follow the pipeline described in \cref{fig:hybrid_representation} are reviewed. In \cite{kumar2009attribute}, low-level local features (image intensities in RGB and HSV colour spaces, edge magnitudes, and gradient directions) were used to compute high-level visual features by training attribute and simile binary SVM classifiers. The attribute classifiers detect describable attributes of faces such as gender, race and age. On the other hand, the simile classifiers detect non-describable attributes by measuring the similarity of different parts of a face to a limited set of reference subjects. To compare two images, the outputs of all the attribute and simile classifiers for both images are fed to an SVM classifier. A method similar to the simile classifiers from \cite{kumar2009attribute} was proposed in \cite{berg2012tom}. The main differences are that \cite{berg2012tom} used a large number of simple one-vs-one classifiers instead of the more complex one-vs-all classifiers used in \cite{kumar2009attribute}, and that SIFT descriptors were used as the low-level features. Two metric learning approaches for face identification were proposed in \cite{guillaumin2009you}. The first one, called \textit{logistic discriminant metric learning} (LDML) is based on the idea that the distance between positive pairs (belonging to the same subject) should be smaller than the distance between negative pairs (belonging to different subjects). The second one, called \textit{marginalised kNN} (MkNN), uses a k-nearest neighbour classifier to find how many positive neighbour pairs can be formed from the neighbours of the two compared vectors. Both methods were trained on pairs of vectors of SIFT descriptors computed at fixed points on the face (corners of the mouth, eyes and nose).

Hybrid methods offer the best of holistic and feature-based methods. Their main limitation is the choice of good features that can fully extract the information needed to recognise a face. Some approaches have tried to overcome this issue by combining different types of features whereas others have introduced a learning stage to improve the discriminative ability of the features. Deep learning methods, discussed next, take these ideas further by training end-to-end systems that can learn a large number of features that are optimal for the recognition task.

\subsection{Deep Learning Methods}
\label{deep_learning_methods}

Convolutional neural networks (CNNs) are the most common type of deep learning method for face recognition. The main advantage of deep learning methods is that they can be trained with large amounts of data to learn a face representation that is robust to the variations present in the training data. In this way, instead of designing specialised features that are robust to different types of intra-class variations (e.g. illumination, pose, facial expression, age, etc.), CNNs can learn them from training data. The main drawback of deep learning methods is that they need to be trained with very large datasets that contain enough variations to generalise to unseen samples. Fortunately, several large-scale face datasets containing in-the-wild face images have recently been released into the public domain \cite{sun2014deep,yi2014learning,parkhi2015deep,guo2016ms,nech2017level,bansal2017umdfaces,cao2017vggface2} to train CNN models. Apart from learning discriminative features, neural networks can reduce dimensionality and be trained as classifiers or using metric learning approaches. CNNs are considered end-to-end trainable systems that do not need to be combined with any other specific methods.

CNN models for face recognition can be trained using different approaches. One of them consists of treating the problem as a classification one, wherein each subject in the training set corresponds to a class. After training, the model can be used to recognise subjects that are not present in the training set by discarding the classification layer and using the features of the previous layer as the face representation \cite{taigman2014deepface}. In the deep learning literature, these features are commonly referred to as \textit{bottleneck} features. Following this first training stage, the model can be further trained using other techniques to optimise the bottleneck features for the target application (e.g. using joint Bayesian \cite{sun2014deep} or fine-tuning the CNN model with a different loss function \cite{yi2014learning}). Another common approach to learning face representation is to directly learn bottleneck features by optimising a distance metric between pairs of faces \cite{chopra2005learning,fan2014learning} or triplets of faces \cite{schroff2015facenet}.

The idea of using neural networks for face recognition is not new. An early method based on a probabilistic decision-based neural network (PBDNN) \cite{lin1997face} was proposed in 1997 for face detection, eye localisation and face recognition. The face recognition PDBNN was divided into one fully-connected subnet per training subject to reduce the number of hidden units and avoid overfitting. Two PBDNNs were trained using intensity and edge features respectively and their outputs were combined to give a final classification decision. Another early method \cite{lawrence1997face} proposed to use a combination of a \textit{self-organising map} (SOM) and a convolutional neural network. A self-organising map \cite{kohonen1998self} is a type of neural network trained in an unsupervised way that projects the input data onto a lower-dimensional space that preserves the topological properties of the input space (i.e. inputs that are nearby in the original space are also nearby in the output space). Note that none of these two early methods were trained end-to-end (edge features were used in \cite{lin1997face} and a SOM in \cite{lawrence1997face}), and that the proposed neural network architectures were shallow. An end-to-end face recognition CNN was proposed in \cite{chopra2005learning}. This method used a \textit{siamese} architecture trained with a contrastive loss function \cite{bromley1994signature}. The contrastive loss implements a metric learning procedure that aims to minimise the distance between pairs of feature vectors corresponding to the same subject while maximising the distance between pairs of feature vectors corresponding to different subjects. The CNN architecture used in this method was also shallow and was trained with small datasets.

None of the methods mentioned above achieved groundbreaking results, mainly due to the low capacity of the networks used and the relatively small datasets available for training at the time. It was not until these models were scaled up and trained with large amounts of data \cite{krizhevsky2012imagenet} that the first deep learning methods for face recognition \cite{taigman2014deepface,sun2014deep} became the state-of-the-art. In particular, Facebook's \textit{DeepFace} \cite{taigman2014deepface}, one of the first CNN-based approaches for face recognition that used a high capacity model, achieved an accuracy of 97.35\% on the LFW benchmark, reducing the error of the previous state-of-the-art by 27\%. The authors trained a CNN with softmax loss\footnote{We refer to \textit{softmax loss} to the combination of the softmax activation function and the cross-entropy loss used to train classifiers.} using a dataset containing 4.4 million faces from 4,030 subjects. Two novel contributions were made in this work: (i) an effective facial alignment system based on explicit 3D modelling of faces, and (ii) a CNN architecture containing locally connected layers \cite{gregor2010emergence,huang2012learning} that (unlike regular convolutional layers) can learn different features from each region in an image. Concurrently, the \textit{DeepID} system \cite{sun2014deep} achieved similar results by training 60 different CNNs on patches comprising ten regions, three scales and RGB or grey channels. During testing, 160 bottleneck features were extracted from each patch and its horizontally flipped counterpart to form a 19,200-dimensional feature vector ($160 \times 2 \times 60$). Similar to \cite{taigman2014deepface}, the proposed CNN architecture also used locally connected layers. The verification result was obtained by training a joint Bayesian classifier \cite{chen2012bayesian} on the 19,200-dimensional feature vectors extracted by the CNNs. The system was trained on a dataset containing 202,599 face images of 10,177 celebrities \cite{sun2014deep}.

\input{tables/tab_datasets}

There are three main factors that affect the accuracy of CNN-based methods for face recognition: training data, CNN architecture, and loss function. As in most deep learning applications, large training sets are needed to prevent overfitting. In general, CNNs trained for classification become more accurate as the number of samples per class increases. This is because the CNN model is able to learn more robust features when is exposed to more intra-class variations. However, in face recognition we are interested in extracting features that generalise to subjects not present in the training set. Hence, the datasets used for face recognition need to also contain a large number of subjects so that the model is exposed to more inter-class variations. The effect that the number of subjects in a dataset has in face recognition accuracy was studied in \cite{zhou2015naive}. In this work, a large dataset was first sorted by the number of images per subject in decreasing order. Then, a CNN was trained with different subsets of training data by gradually increasing the number of subjects. The best accuracy was obtained when the first 10,000 subjects with the most images were used for training. Adding more subjects decreased the accuracy since very few images were available for each extra subject. Another study \cite{bansal2017s} investigated whether wider datasets are better than deeper datasets or vice versa (a dataset is considered wider than another if it contains more subjects; similarly, a dataset is considered deeper than another if it contains more images per subject). From this study, it was concluded that given the same number of images, wider datasets provide better accuracy. The authors argued that this is due to the fact that wider datasets contain more inter-class variations and, therefore, generalise better to unseen subjects. \cref{tab:datasets} shows some of the most common public datasets used to train CNNs for face recognition.

CNN architectures for face recognition have been inspired by those achieving state-of-the-art accuracy on the ImageNet Large Scale Visual Recognition Challenge (ILSVRC). For example, a version of the \textit{VGG} network \cite{simonyan2014very} with 16 layers was used in \cite{parkhi2015deep}, and a similar but smaller network was used in \cite{yi2014learning}. In \cite{schroff2015facenet}, two different types of CNN architectures were explored: VGG style networks \cite{simonyan2014very} and \textit{GoogleNet} style networks \cite{szegedy2015going}. Even though both types of networks achieved comparable accuracy, the GoogleNet style networks had 20 times fewer parameters. More recently, \textit{residual} networks (ResNets) \cite{he2016deep} have become the preferred choice for many object recognition tasks, including face recognition \cite{ranjan2017l2,liu2017sphereface,wu2017deep,hasnat2017deepvisage,wang2018cosface,wang2018additive,deng2018arcface}. The main novelty of ResNets is the introduction of a building block that uses a shortcut connection to learn a residual mapping, as shown in \cref{fig:residual_block}. The use of shortcut connections allows the training of much deeper architectures as they facilitate the flow of information across layers. A thorough study of different CNN architectures was carried out in \cite{deng2018arcface}. The best trade-off between accuracy, speed and model size was obtained with a 100-layer ResNet with a residual block similar to the one proposed in \cite{yamada2016deep}.

\input{figures/fig_residual_block}

The choice of loss function for training CNN-based methods has been the most recent active area of research in face recognition. Even though CNNs trained with softmax loss have been very successful \cite{taigman2014deepface,sun2014deep,yi2014learning,wu2015lightened}, it has been argued that the use of this loss function does not generalise well to subjects not present in the training set. This is because the softmax loss is encouraged to learn features that increase inter-class differences (to be able to separate the classes in the training set) but does not necessarily reduce intra-class variations. Several methods have been proposed to mitigate this issue. A simple approach is to optimise the bottleneck features using a discriminative subspace method such as joint Bayesian \cite{chen2012bayesian}, as done in \cite{sun2014deep,sun2014deep2,wst2008deeply,sun2015deepid3,yi2014learning,chen2016unconstrained}. Another approach is to use metric learning. For example, a pairwise contrastive loss was used as the only supervisory signal in \cite{chopra2005learning,fan2014learning} and combined with a classification loss in \cite{sun2014deep2,wst2008deeply,sun2015deepid3}. One of the most popular metric learning approaches for face recognition is the triplet loss function \cite{weinberger2009distance}, first used in \cite{schroff2015facenet} for the face recognition task. The aim of the triplet loss is to separate the distance between positive pairs from the distance between negative pairs by a margin. More formally, for each triplet $i$ the following condition needs to be satisfied \cite{schroff2015facenet}:
\input{equations/eq_triplet_loss_condition}
where $\bm{x}_a$ is an anchor image, $\bm{x}_p$ is an image of the same subject, $\bm{x}_n$ is an image of a different subject, $f$ is a mapping learnt by a model and $\alpha$ is a margin that is enforced between positive and negative pairs. In practice, CNNs trained with triplet loss converge slower than with softmax loss due to the large number of triplets (or pairs in the case of contrastive loss) needed to cover the entire training set. Although this problem can be alleviated by selecting hard triplets (i.e. triplets that violate the margin condition) during training \cite{schroff2015facenet}, it is common to train with softmax loss in a first training stage and then fine-tune bottleneck features with triplet loss in a second training stage \cite{parkhi2015deep,sankaranarayanan2016triplet,sankaranarayanan2016triplet2}. Some variations of the triplet loss have been proposed. For example, in \cite{sankaranarayanan2016triplet}, the dot product was used as a similarity measure instead of the Euclidean distance; a probabilistic triplet loss was proposed in \cite{sankaranarayanan2016triplet2}; and a modified triplet loss that also minimises the standard deviation of the distributions of positive and negative scores was proposed in \cite{kumar2016learning,trigueros2018enhancing}. An alternative loss function used to learn discriminative features is the centre loss proposed in \cite{wen2016discriminative}. The goal of the centre loss is to minimise the distances between bottleneck features and their corresponding class centres. By jointly training with softmax and centre loss, it was shown that the features learnt by a CNN could effectively increase inter-personal variations (softmax loss) and reduce intra-personal variations (centre loss). The centre loss has the advantage of being more efficient and easier to implement than the contrastive and triplet losses since it does not require forming pairs or triplets during training. Another related metric learning method is the range loss proposed in \cite{zhang2016range} for improving training with unbalanced datasets. The range loss has two components. The intra-class component of the loss minimises the \textit{k}-largest distances between samples of the same class, and the inter-class component of the loss maximises the distance between the closest two class centres in each training batch. By using these extreme cases, the range loss uses the same information from each class, regardless of how many samples per class are available. Similar to the centre loss, the range loss needs to be combined with softmax loss to avoid the loss being degraded to zeros \cite{wen2016discriminative}.

\input{figures/fig_decision_boundaries}

One of the difficulties that arise when combining different loss functions is finding the correct balance between each term. Recently, several approaches have proposed to modify the softmax loss so that it can learn discriminative features with no need to combine it with other losses. One approach that has been shown to increase the discriminative ability of the bottleneck features is feature normalisation \cite{ranjan2017l2,hasnat2017deepvisage}. For example, \cite{ranjan2017l2} proposed to normalise the features to have unit $L_2$-norm and \cite{hasnat2017deepvisage} proposed to normalise the features to have zero mean and unit variance. A very successful development has been the introduction of a margin in the decision boundary between each class in the softmax loss \cite{liu2016large}. For simplicity, consider binary classification with softmax loss. In this case, the decision boundary between each class (if the biases are zero) is given by:
\input{equations/eq_softmax_decision_boundary}
where $\bm{x}$ is a feature vector, $\bm{W}_1$ and $\bm{W}_2$ are the weights corresponding to each class and $\theta_1$ and $\theta_2$ are the angles between $\bm{x}$ and $\bm{W}_1$ and $\bm{W}_2$ respectively. By introducing a multiplicative margin $m$ in \cref{eq:softmax_decision_boundary}, the two decision boundaries become more stringent:
\input{equations/eq_softmax_margin_decision_boundary}
As shown in \cref{fig:decision_boundaries}, the margin can effectively increase the separation between classes and their intra-class compactness. Several alternative approaches have been proposed depending on how the margin is incorporated into the loss \cite{liu2017sphereface,wang2018cosface,wang2018additive,deng2018arcface}. For example, in \cite{liu2017sphereface} the weight vectors were normalised to have unit norm so that the decision boundary only depends on the angles $\theta_1$ and $\theta_2$. In \cite{wang2018cosface,wang2018additive}, an additive cosine margin was proposed. Compared to the multiplicative margin \cite{liu2016large,liu2017sphereface}, the additive margin is easier to implement and optimise. In this work, apart from normalising the weight vectors, the feature vectors were also normalised and scaled as done in \cite{ranjan2017l2}. An alternative additive margin was proposed in \cite{deng2018arcface} which keeps the advantages of \cite{wang2018cosface,wang2018additive} but has a better geometric interpretation since the margin is added to the angle and not to the cosine. \cref{tab:decision_boundaries} summarises the decision boundaries for the different variations of the softmax loss with margin. These approaches are the current state-of-the-art in face recognition.

\input{tables/tab_decision_boundaries}

\section{Conclusions}
\label{conclusions}

We have seen how face recognition has followed the same transition as many other computer vision applications. Traditional methods based on hand-engineered features that provided state-of-the-art accuracy only a few years ago have been replaced by deep learning methods based on CNNs. Indeed, face recognition systems based on CNNs have become the standard due to the significant accuracy improvement achieved over other types of methods. Moreover, it is straightforward to scale-up these systems to achieve even higher accuracy by increasing the size of the training sets and/or the capacity of the networks. However, collecting large amounts of labelled face images is expensive, and very deep CNN architectures are slow to train and deploy. Generative adversarial networks (GANs) \cite{goodfellow2014generative} are a promising solution to the first issue. Recent works on GANs with face images include facial attributes manipulation \cite{larsen2015autoencoding,perarnau2016invertible,brock2016neural,shen2017learning,lu2017conditional,choi2017stargan,yin2017semi,shu2017neural,lample2017fader,he2017arbitrary}, facial expression editing \cite{zhou2017photorealistic,ding2017exprgan,choi2017stargan}, generation of novel identities \cite{donahue2017semantically}, face frontalisation \cite{huang2017beyond,tran2017representation} and face ageing \cite{antipov2017face,zhang2017age}. It is expected that these advancements will be used to generate additional training images without requiring millions of face images to be labelled. To address the second issue, more efficient architectures such as MobileNets \cite{howard2017mobilenets,sandler2018inverted} are being developed and used for real-time face recognition on devices with limited computational resources \cite{chen2018mobilefacenets}.

\bibliographystyle{ieeetr}
\bibliography{bibliography}

\end{document}

%% file: figures/fig_face_variations.tex
\begin{figure}[tb]
    \centering
    \subcaptionbox{\label{fig:face_variations_a}}{
        \includegraphics[width=0.16\linewidth]{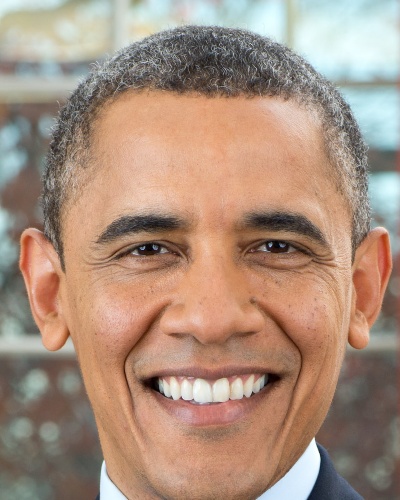}
        \includegraphics[width=0.16\linewidth]{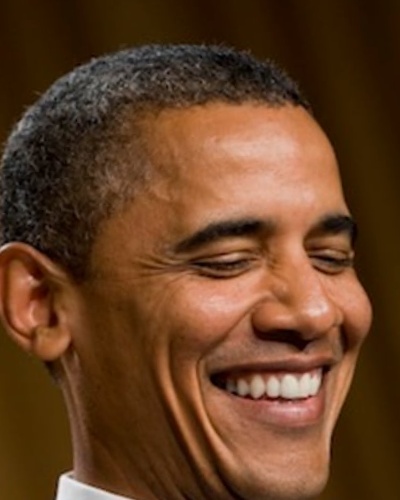}
    }\hspace{0.038\linewidth}
    \subcaptionbox{\label{fig:face_variations_b}}{
        \includegraphics[width=0.16\linewidth]{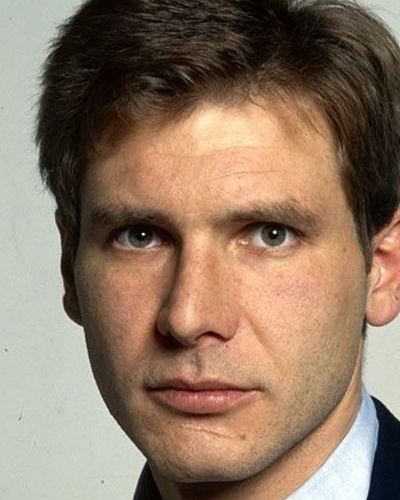}
        \includegraphics[width=0.16\linewidth]{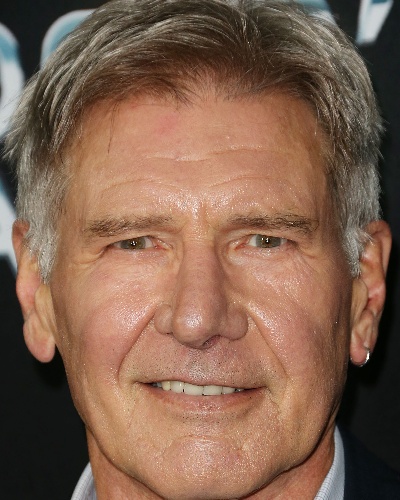}
    }
    \par\medskip
    \subcaptionbox{\label{fig:face_variations_c}}{
        \includegraphics[width=0.16\linewidth]{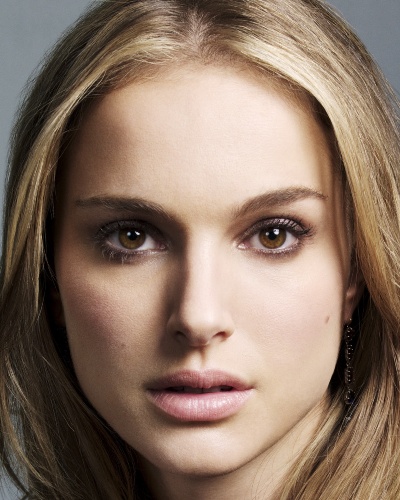}
        \includegraphics[width=0.16\linewidth]{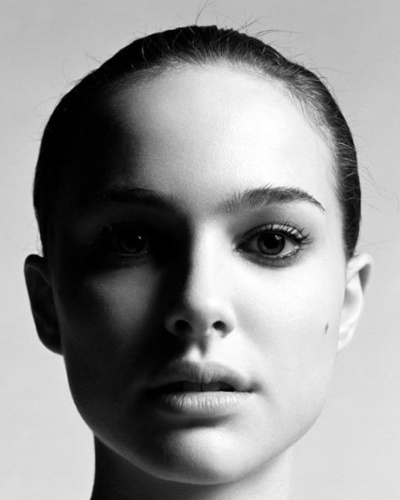}
    }\hspace{0.038\linewidth}
    \subcaptionbox{\label{fig:face_variations_d}}{
        \includegraphics[width=0.16\linewidth]{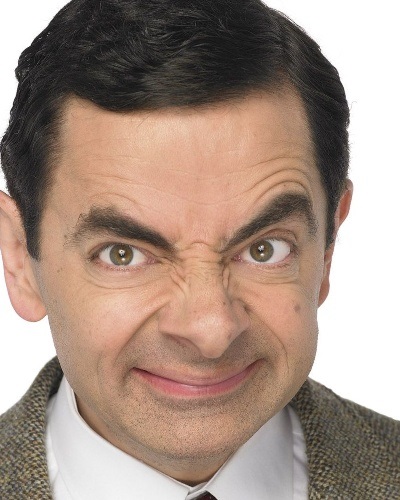}
        \includegraphics[width=0.16\linewidth]{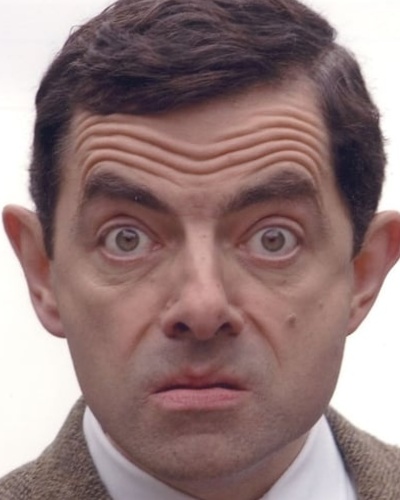}
    }\hspace{0.038\linewidth}
    \subcaptionbox{\label{fig:face_variations_e}}{
        \includegraphics[width=0.16\linewidth]{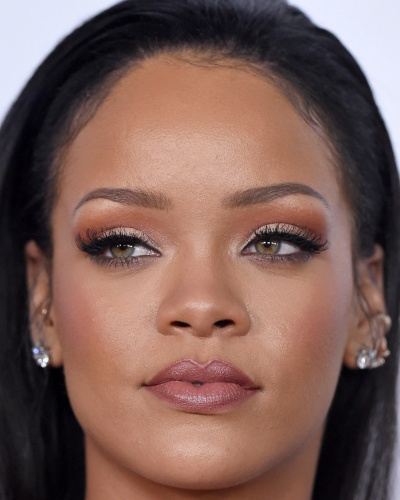}
        \includegraphics[width=0.16\linewidth]{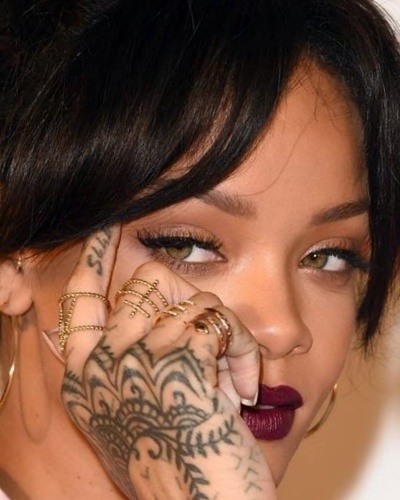}
    }
    \caption{Typical variations found in faces in-the-wild. \subref*{fig:face_variations_a} Head pose. \subref*{fig:face_variations_b} Age. \subref*{fig:face_variations_c} Illumination. \subref*{fig:face_variations_d} Facial expression. \subref*{fig:face_variations_e} Occlusion.}
    \label{fig:face_variations}
\end{figure}

%% file: figures/fig_face_recognition_blocks.tex
\begin{figure*}[tb]
    \centering
    \includegraphics[width=0.85\linewidth]{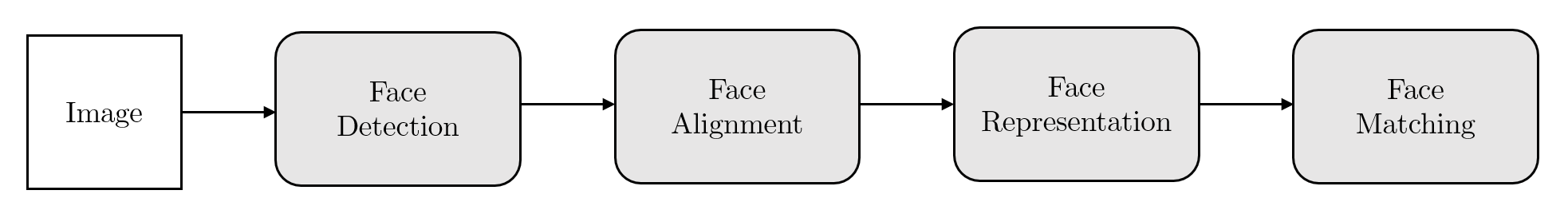}
    \caption{Face recognition building blocks.}
    \label{fig:face_recognition_blocks}
\end{figure*}

%% file: figures/fig_face_detection_alignment.tex
\newsavebox{\largestimage}
\begin{figure}[b]
    \centering
    \savebox{\largestimage}{\includegraphics[width=0.5\linewidth]{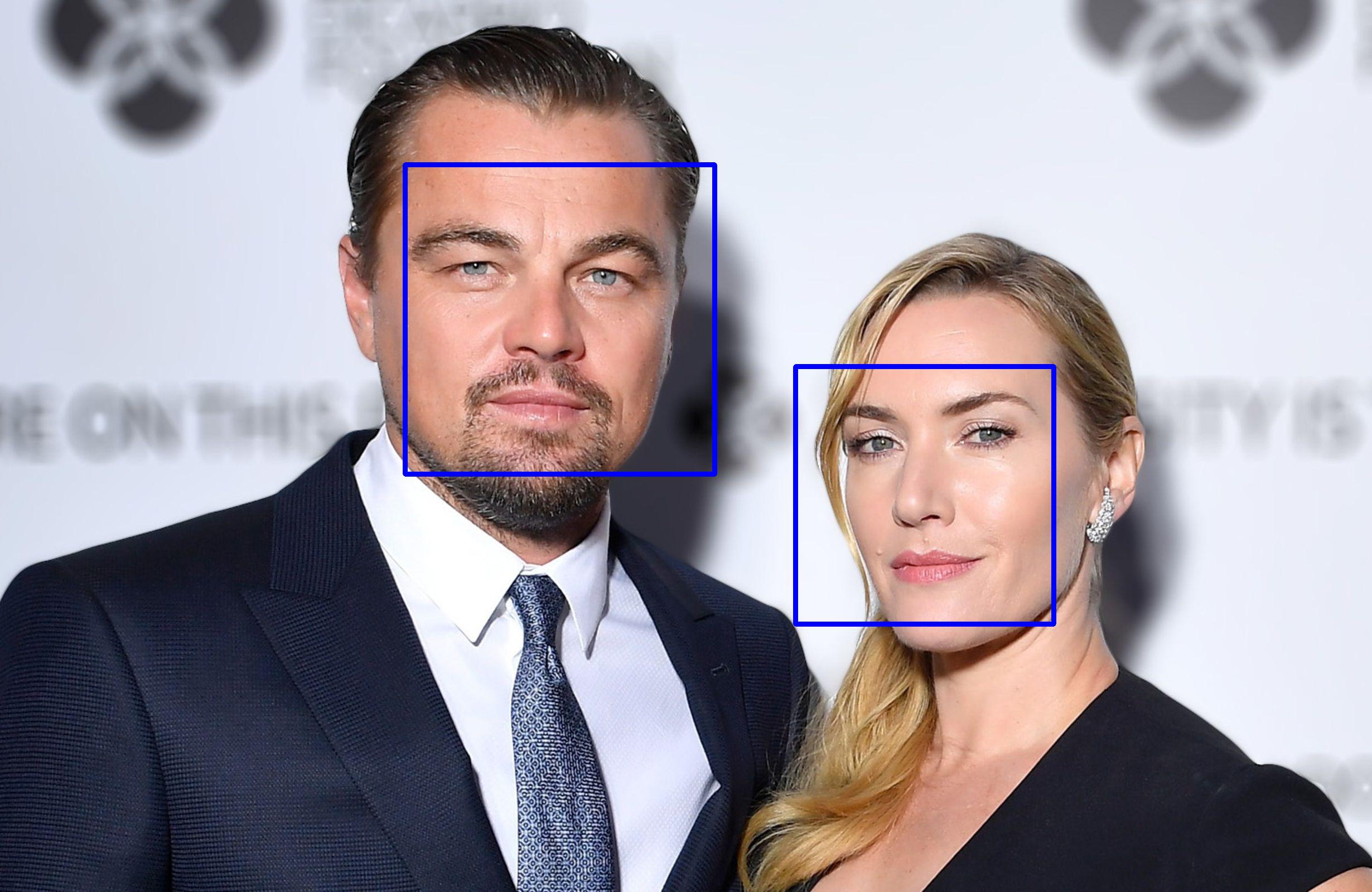}}
    \hspace*{\fill}
    \subcaptionbox{\label{fig:face_detection_alignment_a}}{\usebox{\largestimage}}
    \hspace*{\fill}
    \subcaptionbox{\label{fig:face_detection_alignment_b}}{\raisebox{\dimexpr.5\ht\largestimage-.5\height}{\includegraphics[width=0.2\linewidth]{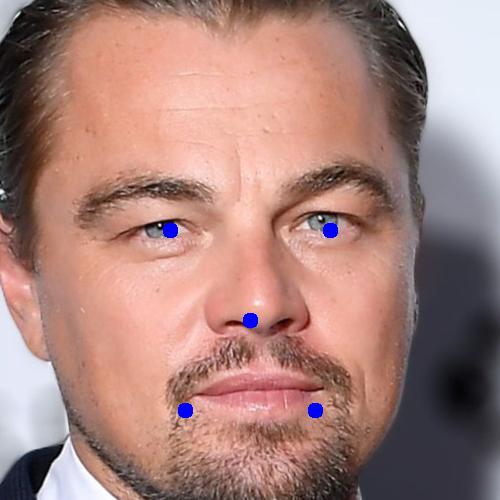}}}
    \subcaptionbox{\label{fig:face_detection_alignment_c}}{\raisebox{\dimexpr.5\ht\largestimage-.5\height}{\includegraphics[width=0.2\linewidth]{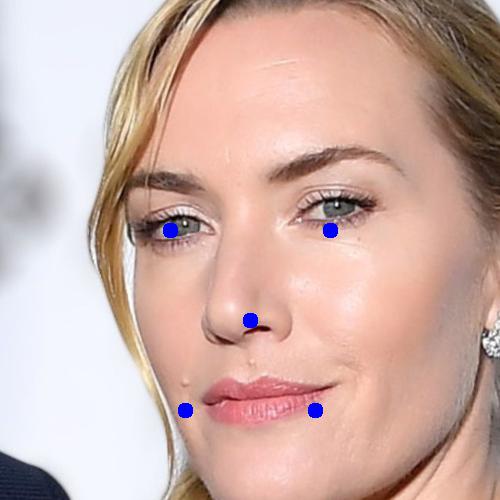}}}
    \hspace*{\fill}
    \caption{\subref*{fig:face_detection_alignment_a} Bounding boxes found by a face detector. \subref*{fig:face_detection_alignment_b} and \subref*{fig:face_detection_alignment_c} Aligned faces and reference points.}
    \label{fig:face_detection_alignment}
\end{figure}

%% file: figures/fig_eigenfaces.tex
\begin{figure}[tb]
    \centering
    \includegraphics[width=0.15\linewidth]{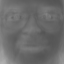}\hspace{0.04\linewidth}
    \includegraphics[width=0.15\linewidth]{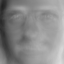}\hspace{0.04\linewidth}
    \includegraphics[width=0.15\linewidth]{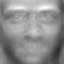}\hspace{0.04\linewidth}
    \includegraphics[width=0.15\linewidth]{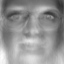}\hspace{0.04\linewidth}
    \includegraphics[width=0.15\linewidth]{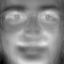}
    \caption{Top 5 eigenfaces computed using the ORL database of faces \cite{samaria1994parameterisation} sorted from most variance (left) to least variance (right).}
    \label{fig:eigenfaces}
\end{figure}

%% file: equations/eq_lda_projection_matrix.tex
\begin{equation}
    \bm{W}^*=\argmax_{\bm{W}}\frac{|\bm{W}^T\bm{S}_b\bm{W}|}{|\bm{W}^T\bm{S}_w\bm{W}|}
    \label{eq:lda_projection_matrix}
\end{equation}

%% file: equations/eq_lda_scatter_matrices.tex
\begin{align}
    \bm{S}_w &= \sum^K_k{\sum_{\bm{x}_j \in C_k}{(\bm{x}_j-\bm{\mu}_k)(\bm{x}_j-\bm{\mu}_k)^T}}
    \label{eq:lda_between_scatter_matrix} \\
    \bm{S}_b &= \sum^K_k{(\bm{\mu}-\bm{\mu}_k)(\bm{\mu}-\bm{\mu}_k)^T}
    \label{eq:lda_whithin_scatter_matrix}
\end{align}

%% file: equations/eq_src_representation.tex
\begin{equation}
    \bm{y} = \bm{A}\bm{x}_0
    \label{eq:src_representation}
\end{equation}

%% file: equations/eq_src_representation_corrupted.tex
\begin{equation}
    \bm{y} = \bm{A}\bm{x}_0+\bm{e}_0
    \label{eq:src_representation_corrupted}
\end{equation}

%% file: equations/eq_chi_square_distance.tex
\begin{equation}
    \chi^2(\bm{a}, \bm{b}) = \sum_i{\frac{w_i(a_i-b_i)^2}{a_i+b_i}}
    \label{eq:chi_square_distance}
\end{equation}

%% file: figures/fig_lbp_representation.tex
\begin{figure}[tb]
    \centering
    \subcaptionbox{\label{fig:lbp_representation_a}}{\includegraphics[width=0.4\linewidth]{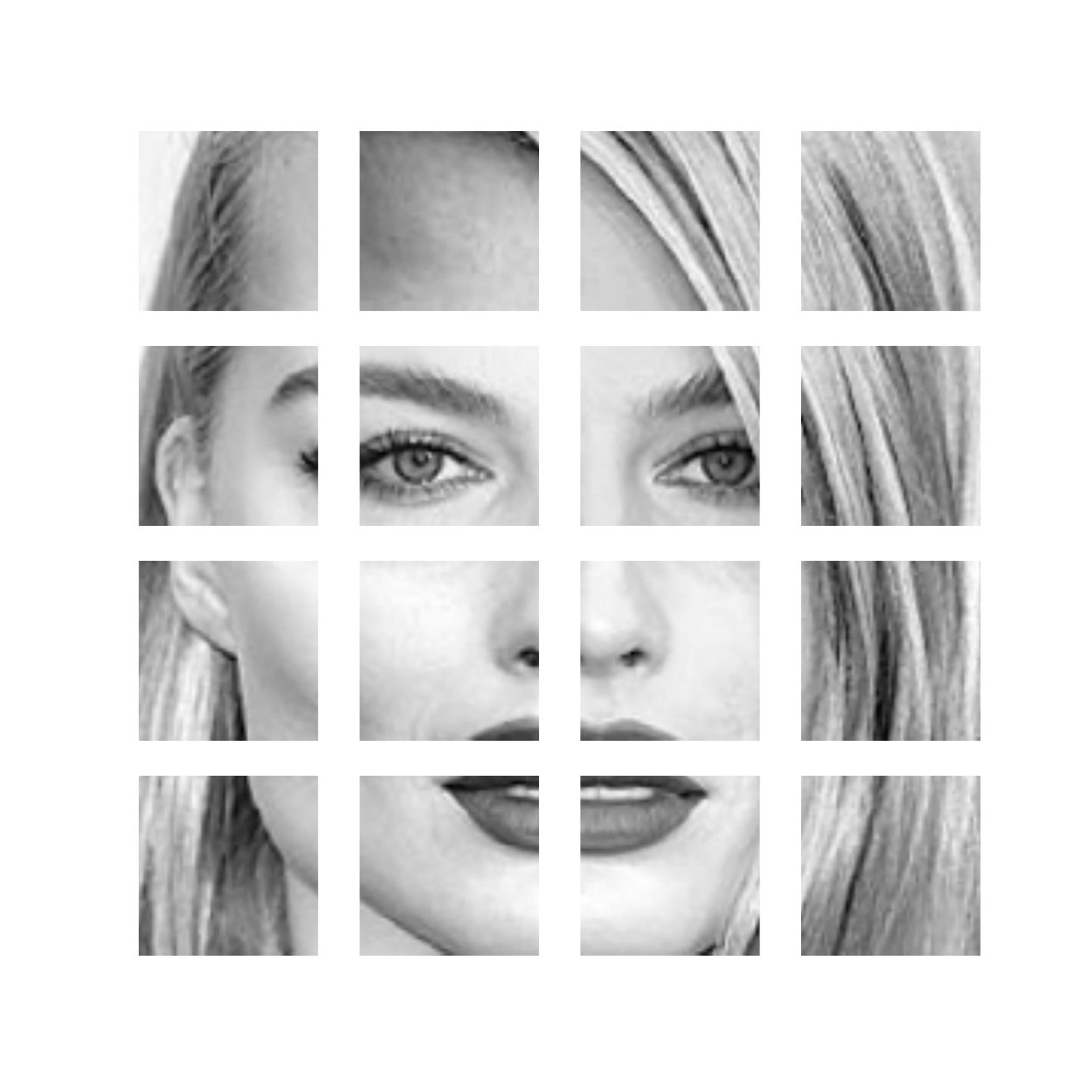}}
    \subcaptionbox{\label{fig:lbp_representation_b}}{\includegraphics[width=0.4\linewidth]{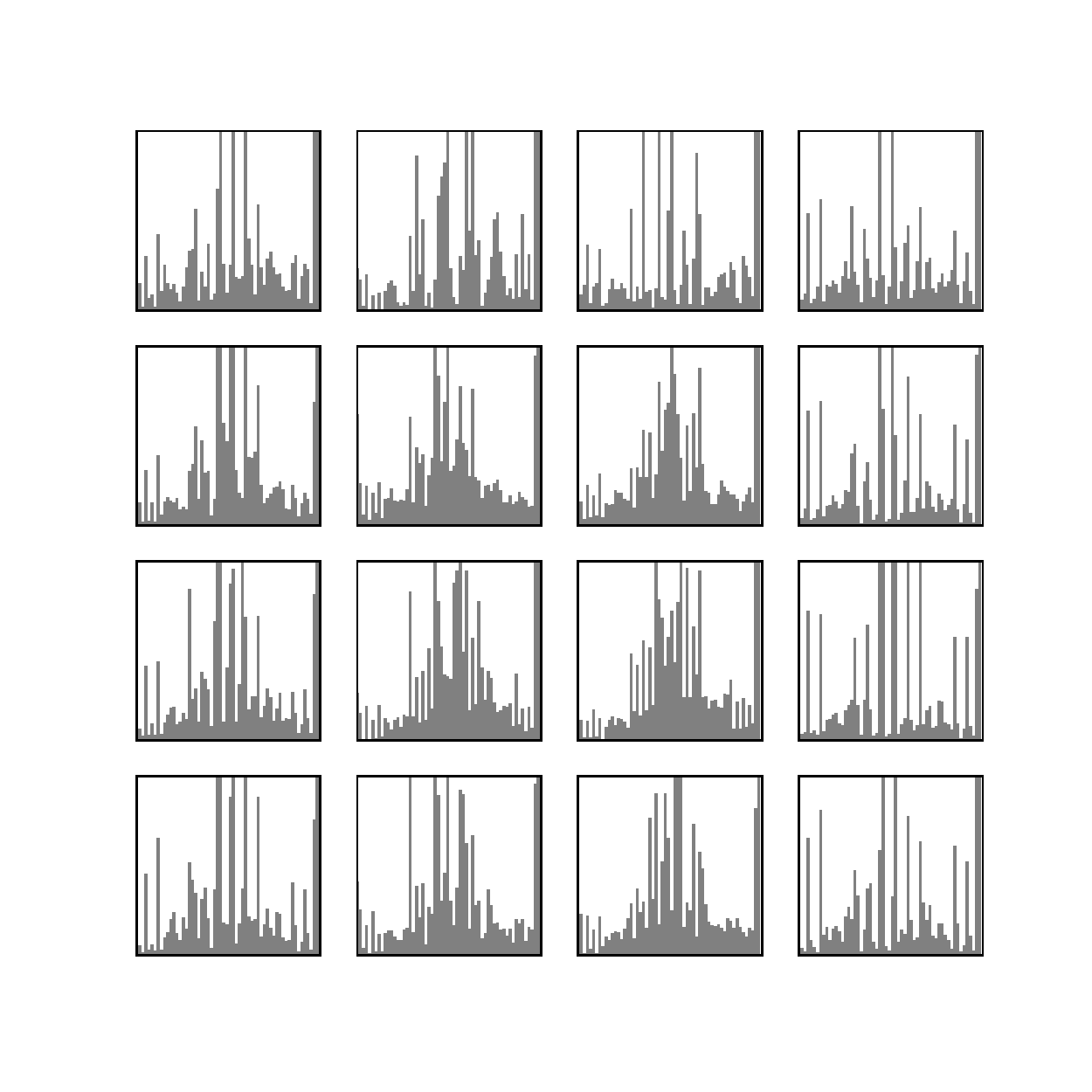}}
    \caption{\subref*{fig:lbp_representation_a} Face image divided into $4 \times 4$ local regions. \subref*{fig:lbp_representation_b} Histograms of LBP descriptors computed from each local region.}
    \label{fig:lbp_representation}
\end{figure}

%% file: equations/eq_sparse_projection.tex
\begin{equation}
    \min_{\bm{B}}\|\bm{Y}-\bm{B}^T\bm{X}\|_2^2+\lambda\|\bm{B}\|_1
    \label{eq:sparse_projection}
\end{equation}

%% file: figures/fig_hybrid_representation.tex
\begin{figure}[tb]
    \centering
    \includegraphics[width=0.7\linewidth]{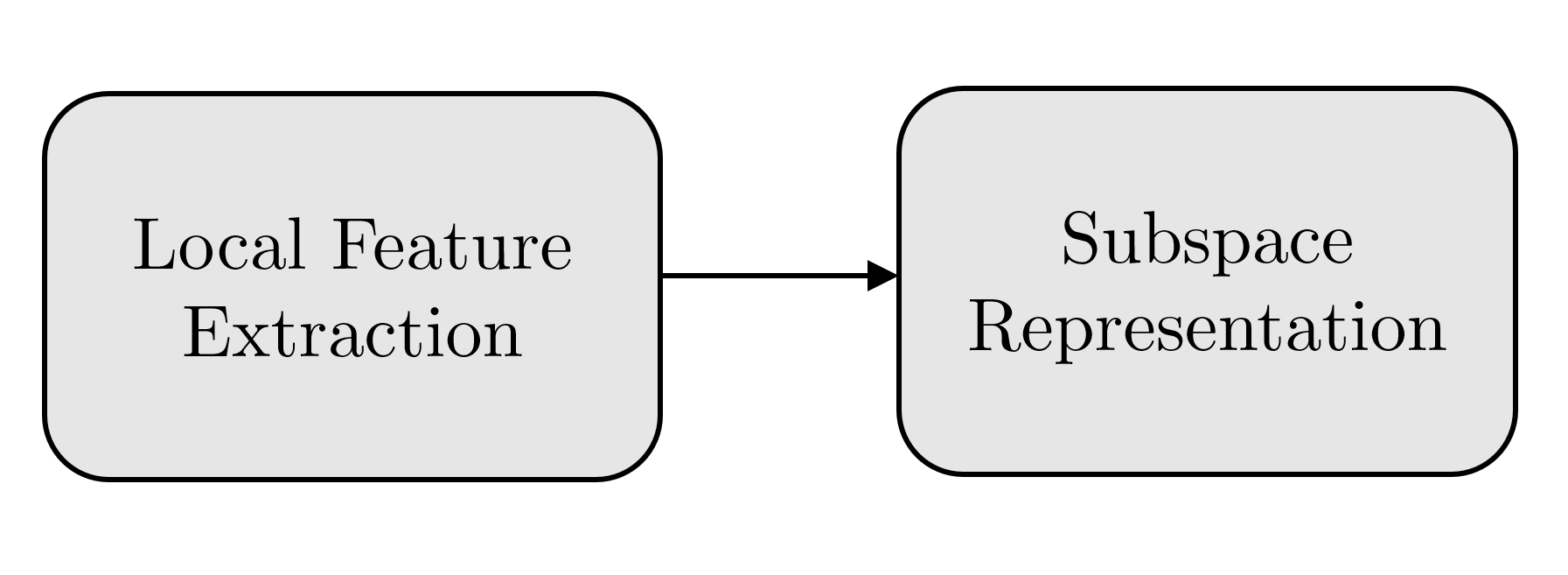}
    \caption{Typical hybrid face representation.}
    \label{fig:hybrid_representation}
\end{figure}

%% file: tables/tab_datasets.tex
\begin{table}[tb]
    \centering
    \caption{Public large-scale face datasets.}
    \label{tab:datasets}
    \begin{tabular}{@{}lccc@{}}
    \toprule
    Dataset                             & Images           & Subjects           & Images per subject \\ \midrule
    CelebFaces+ \cite{sun2014deep}      & 202,599          & 10,177             & 19.9               \\
    UMDFaces \cite{bansal2017umdfaces}  & 367,920          & 8,501              & 43.3               \\
    CASIA-WebFace \cite{yi2014learning} & 494,414          & 10,575             & 46.8               \\
    VGGFace \cite{parkhi2015deep}       & 2.6M             & 2,622              & 1,000              \\
    VGGFace2 \cite{cao2017vggface2}     & 3.31M            & 9,131              & 362.6              \\
    MegaFace \cite{nech2017level}       & 4.7M             & 672,057            & 7                  \\
    MS-Celeb-1M \cite{guo2016ms}        & 10M              & 100,000            & 100                \\ \bottomrule
    \end{tabular}
\end{table}

%% file: figures/fig_residual_block.tex
\begin{figure}[tb]
    \centering
    \includegraphics[width=0.5\linewidth]{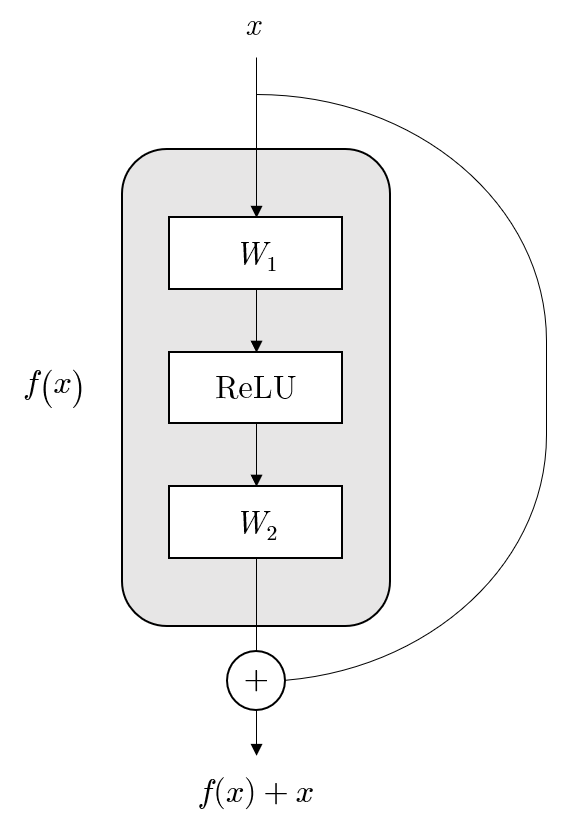}
    \caption{Original residual block proposed in \cite{he2016deep}.}
    \label{fig:residual_block}
\end{figure}

%% file: equations/eq_triplet_loss_condition.tex
\begin{equation}
    \norm{f(\bm{x}_a)-f(\bm{x}_p)}_2^2 + \alpha < \norm{f(\bm{x}_a)-f(\bm{x}_n)}_2^2\label{eq:triplet_loss_condition}
\end{equation}

%% file: figures/fig_decision_boundaries.tex
\begin{figure}[tb]
    \centering
    \subcaptionbox{\label{fig:decision_boundary_a}}{\includegraphics[width=0.45\linewidth]{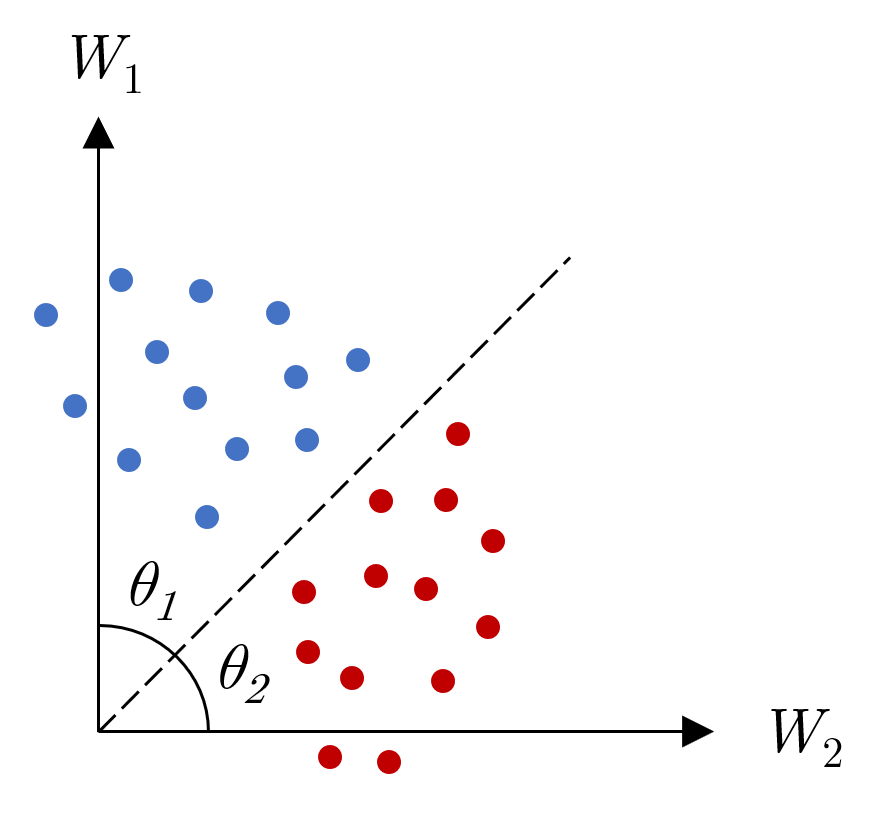}}\hspace{0.05\linewidth}
    \subcaptionbox{\label{fig:decision_boundary_b}}{\includegraphics[width=0.45\linewidth]{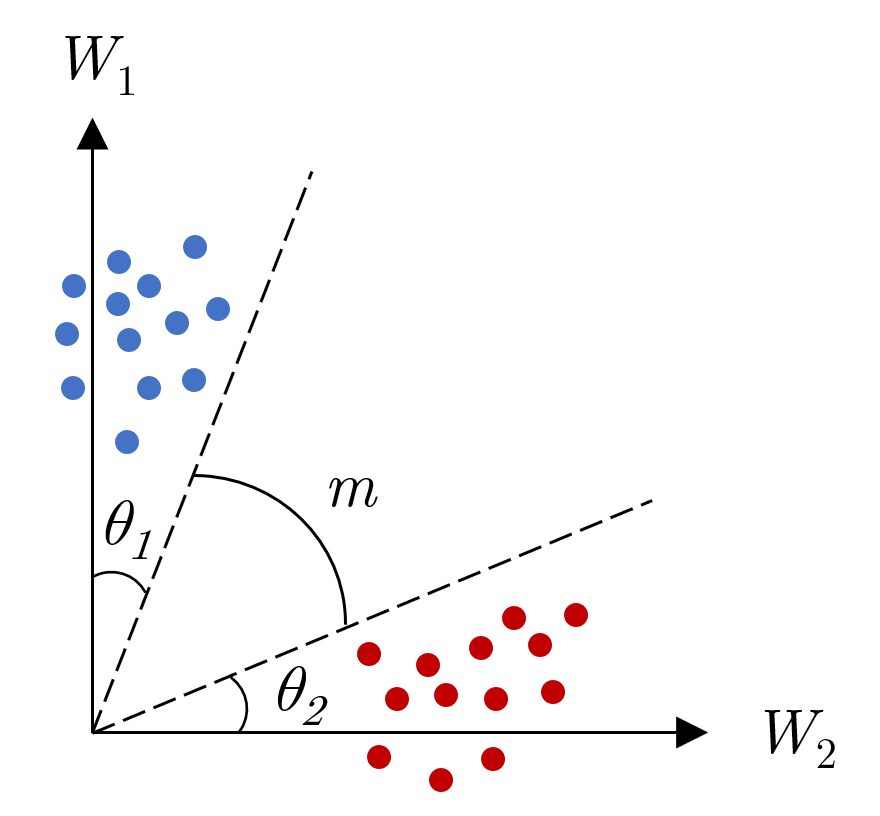}}
    \caption{Effect of introducing a margin $m$ in the decision boundary between two classes. \subref*{fig:decision_boundary_a} Softmax loss. \subref*{fig:decision_boundary_b} Softmax loss with margin.}
    \label{fig:decision_boundaries}
\end{figure}

%% file: equations/eq_softmax_decision_boundary.tex
\begin{equation}
    \norm{\bm{x}}(\norm{\bm{W}_1}\cos\theta_1-\norm{\bm{W}_2}\cos\theta_2)=0\label{eq:softmax_decision_boundary}
\end{equation}

%% file: equations/eq_softmax_margin_decision_boundary.tex
\begin{align}
    \norm{\bm{x}}(\norm{\bm{W}_1}\cos m\theta_1-\norm{\bm{W}_2}\cos\theta_2)=0 \text{ for class 1}\label{eq:softmax_margin_decision_boundary_1} \\
    \norm{\bm{x}}(\norm{\bm{W}_1}\cos\theta_1-\norm{\bm{W}_2}\cos m\theta_2)=0 \text{ for class 2}\label{eq:softmax_margin_decision_boundary_2}
\end{align}

%% file: tables/tab_decision_boundaries.tex
\begin{table}[tb]
    \centering
    \caption{Decision boundaries for different variations of the softmax loss with margin. Note that the decision boundaries are for class 1 in a binary classification case.}
    \label{tab:decision_boundaries}
    \begin{tabular}{@{}ll@{}}
    \toprule
    Type of softmax margin                                         & Decision boundary                              \\ \midrule
    Multiplicative angular margin \cite{liu2017sphereface}         & $\norm{\bm{x}}(\cos m\theta_1-\cos\theta_2)=0$ \\
    Additive cosine margin \cite{wang2018cosface,wang2018additive} & $s(\cos\theta_1-m-\cos\theta_2)=0$             \\
    Additive angular margin \cite{deng2018arcface}                 & $s(\cos(\theta_1+m)-\cos\theta_2)=0$           \\ \bottomrule
    \end{tabular}
\end{table}